\documentclass[lettersize,journal]{IEEEtran}
\usepackage{amsmath,amsfonts}
\usepackage{algorithmic}
\usepackage{array}
\usepackage{textcomp}
\usepackage{url}
\usepackage{verbatim}
\usepackage{graphicx}
\usepackage{hyperref}
\hypersetup{
    colorlinks=true,
    linkcolor=red,
    citecolor=blue,
    urlcolor=blue
}

\usepackage{cleveref}
\usepackage{color}
\usepackage{xcolor}
\usepackage{multirow}
\usepackage{colortbl}
\usepackage{booktabs}
\newcommand{\newcref}[1]{\textcolor{red}{\Cref{#1}}}

\crefname{figure}{Fig.}{Figs.}
\crefname{table}{Tab.}{Tabs.}
\crefname{section}{Sec.}{Secs.}
\crefname{equation}{Eq.}{Eqs.}

\begin{document}

\title{SD-ReID: View-aware Stable Diffusion for Aerial-Ground Person Re-Identification}

\author{Yuhao Wang, Xiang Hu, Lixin Wang, Pingping Zhang,\IEEEmembership{~IEEE Member}, Huchuan Lu,\IEEEmembership{~IEEE Fellow}
\thanks{Copyright (c) 2026 IEEE. Personal use of this material is permitted. However, permission to use this material for any other purposes must  be obtained from the IEEE by sending an email to \textcolor{blue}{\underline{pubs-permissions@ieee.org}}.

%
(Corresponding author: Pingping Zhang.)}

\thanks{Yuhao Wang, Xiang Hu, Lixin Wang and Pingping Zhang are with the School of Future Technology, Dalian University of Technology, Dalian, 116024, China. (Email: 924973292@mail.dlut.edu.cn; 1908414518@mail.dlut.edu.cn;wanglixin@mail.dlut.edu.cn;zhpp@dlut.edu.cn)}

\thanks{Huchuan Lu is with the School of Information and Communication Engineering, Dalian University of Technology, Dalian, 116024, China. (Email: lhchuan@dlut.edu.cn)}
}

\markboth{IEEE Transactions on Image Processing}{}

\maketitle
\begin{abstract}
Aerial-Ground Person Re-IDentification (AG-ReID) aims to retrieve specific persons across cameras with different viewpoints.
Previous works focus on designing discriminative models to maintain the identity consistency despite drastic changes in camera viewpoints.
The core idea behind these methods is quite natural, but designing a view-robust model is a very challenging task.
Moreover, they overlook the contribution of view-specific features in enhancing the model's ability to represent persons.
To address these issues, we propose a novel generative framework named SD-ReID for AG-ReID, which leverages generative models to mimic the feature distribution of different views while extracting robust identity representations.
More specifically, we first train a ViT-based model to extract person representations along with controllable conditions, including identity and view conditions.
We then fine-tune the Stable Diffusion (SD) model to enhance person representations guided by these controllable conditions.
Furthermore, we introduce the View-Refined Decoder (VRD) to bridge the gap between instance-level and global-level features.
Finally, both person representations and all-view features are employed to retrieve target persons.
Extensive experiments on five AG-ReID benchmarks (i.e., CARGO, AG-ReIDv1, AG-ReIDv2, LAGPeR and G2APS-ReID) demonstrate the effectiveness of our proposed method.
The source code and pre-trained models are available at https://github.com/924973292/SD-ReID.
\end{abstract}

\section{Introduction}
\label{sec:intro}
Person Re-IDentification (ReID) aims to retrieve the same person across non-overlapping cameras.
In recent years, ReID has gained much attention due to the urgent demand of societal security, intelligent surveillance, mobile robotics and human-computer interaction.
Many methods have been proposed to address the challenges in ReID, such as illumination variation~\cite{huang2019illumination}, low-image resolution~\cite{li2015multi} and occlusion~\cite{huang2018adversarially}.
However, these methods are primarily based on datasets collected from fixed ground cameras or CCTV systems.
This characteristic makes the practical performance of these methods highly dependent on the density of camera deployment.
However, the deployment of fixed cameras is often limited by complex factors, such as environmental condition and infrastructure availability.
Thus, it suffers from sparse camera coverage, which poses challenges for reliably retrieving target persons.
\begin{figure}[t]
\centering
\includegraphics[width=1.00\linewidth]{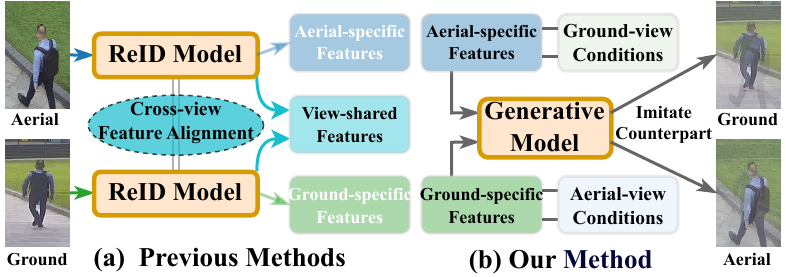}
\caption{Motivations. (a) Previous AG-ReID methods focus on extracting view-shared features through cross-view feature alignment while discarding view-specific ones.
(b) Our method leverages view-specific features with generative models conditioned on the opposite view to imitate counterparts.}
\label{fig:moti}
\vspace{-2mm}
\end{figure}

With the rapid development of Unmanned Aerial Vehicles (UAVs), deploying cameras on drones has become increasingly prevalent.
Consequently, Aerial-Ground person ReID (AG-ReID), which addresses the substantial cross-platform discrepancies between aerial and ground perspectives, has attracted significant attention~\cite{nguyen2023aerial,zhang2024view,zhou2025text}.
The core challenge of AG-ReID lies in the drastic visual variations caused by platform-dependent viewpoint changes.
Recently, several works have attempted to address this issue.
For example, Zhang et al.~\cite{zhang2024view} disentangle identity and view information from input images.
Meanwhile, Nguyen et al.~\cite{nguyen2024ag} propose a two-stream explainable model to exploit person attributes and enhance feature learning.
However, as shown in \newcref{fig:moti} (a), previous works focus on extracting view-shared features by aligning cross-view features while discarding view-specific ones.
In practice, view-shared features are difficult to extract due to various challenges, particularly the drastic view changes in AG-ReID.
Furthermore, the positive contribution of view-specific features for cross-view retrieval has been largely overlooked.
As illustrated in \newcref{fig:moti} (b), our key motivation is to leverage these view-specific features using generative models, enabling each view to imitate the feature distribution of its counterpart.
This approach not only preserves the unique information of each view but also enhances cross-view retrieval in a more flexible and adaptive manner.
Meanwhile, with the development of large-scale pre-trained generative models, e.g., Stable Diffusion (SD)~\cite{rombach2022high}, applying diffusion models to discriminative tasks has emerged as an outstanding approach for addressing domain-specific challenges~\cite{chen2023diffusiondet,chen2023generalist}.
Despite this progress, existing generative approaches in ReID~\cite{chen2023generalist,kim2024pose} employ diffusion models solely for training-time data augmentation.
They discard the generative model during inference, leaving its capacity unexploited at retrieval time.
Going beyond this paradigm, we integrate diffusion models into AG-ReID to synthesize complementary features conditioned on the opposite view, thereby enhancing cross-view retrieval.

Motivated by these observations, we propose a novel two-stage framework named SD-ReID for AG-ReID.
Technically, in the first stage, we train a simple view-aware ReID model to coarsely extract person representations along with conditions encoding identity and view information.
In the second stage, all parameters of the ReID model are fixed.
Then, we use the person representations obtained from the first stage as generation targets to train the SD model.
To improve the controllability of the generation process, we design a condition learner that injects identity and view conditions into the SD model.
With this design, the model can generate view-specific features.
While this mechanism works effectively during training, instance-level view conditions cannot be accessed during inference, since the image of the same person with other views is unavailable in real retrieval scenarios.
To address this limitation, we construct a memory bank to aggregate global-level view conditions, which are obtained by averaging view representations across the training set.
These global-level conditions provide approximate guidance for generation at inference.
However, they inevitably introduce a distribution gap compared to instance-level conditions, which may degrade the quality and discrimination of generated features.
To mitigate this issue, we propose the View-Refined Decoder (VRD) to adaptively refine generated features by aligning them with visual features from the ReID backbone.
Finally, the refined all-view features generated by the SD model are fused with visual features extracted by the ReID model, enabling a more robust and comprehensive representation for person retrieval.

In summary, our key contributions are as follows:
\begin{itemize}
    \item
    To the best of our knowledge, we are the first to introduce generative models into AG-ReID to directly synthesize person representations with cross-view information.
    \item
    We introduce a novel feature learning framework named SD-ReID for AG-ReID, which enhances view-invariant representations by imitating view-specific feature distributions with generative models.
    \item
    We design the View-Refined Decoder (VRD) to bridge the gap between instance-level and global-level view conditions, thereby improving the quality and discrimination of generated features for better cross-view retrieval.
    \item
    Extensive experiments on five AG-ReID benchmarks validate the effectiveness of our proposed SD-ReID.
\end{itemize}
\section{Related Work}
\label{sec:Rela}
\subsection{Image-based Person ReID}
Image-based person ReID has been studied as a fundamental task in intelligent security systems.
Early approaches mainly relied on handcrafted features.
With the advent of deep neural networks, Convolutional Neural Networks (CNNs) are introduced to enhance feature representation~\cite{yi2014deep,zhou2019omni,quan2019auto}.
However, CNNs suffer from limited receptive fields, which restrict their ability to model long-range dependencies in complex scenes.
This limitation has motivated the development of Transformer-based methods~\cite{he2021transreid,zhang2024view,li2023clip,zhou2026hierarchical, zhang2025dual, li2025breaking,zhang2025weakly}, which leverage attention mechanisms to achieve more robust representations.
Despite these advances, most existing methods are designed for single-type camera inputs and still struggle in complex scenarios due to the inherent constraints of camera systems.
To alleviate these problems, cross-modality and multi-modality ReID tasks have been introduced, such as visible-thermal ReID~\cite{wu2017rgb,ye2018hierarchical,lu2023learning,yu2026x} and multi-modal ReID~\cite{wang2024top,zhang2024magic,wang2025mambapro,wang2025decoupled,wang2025idea,xu2026stmi,liu2026signal}, enabling retrieval across heterogeneous camera types.
However, these approaches typically rely on fixed surveillance cameras, which only capture limited viewpoints and perform poorly in sparse-camera scenarios.
This shortcoming has motivated the emergence of AGReID, a more challenging setting that requires bridging drastic cross-platform viewpoint discrepancies while ensuring reliable performance in real-world applications.
\subsection{Aerial-Ground Person ReID}
AG-ReID has attracted increasing attention due to its practical relevance in scenarios with sparse camera coverage.
Nguyen et al.~\cite{nguyen2023aerial} first introduce the AG-ReID v1 dataset and propose a two-stream explainable model that leverages person attributes to identify individuals across aerial and ground views.
Nguyen et al.~\cite{nguyen2024ag} further extend this dataset to AG-ReID v2 and develop a local stream to extract head-region features.
Zhang et al.~\cite{zhang2024view} construct the synthetic CARGO dataset and employ a view token to hierarchically disentangle view-invariant features.
Wang et al.~\cite{wang2025secap} utilize view-aware prompts to decode local invariant representations and further introduce the LAGPeR and G2APS-ReID datasets.
More recently, several works~\cite{zhang2024cross,hambarde2025detreidx,ha2025multi,nguyen2025ag,nguyen2025agvir} extend AG-ReID to video-based and multi-modality scenarios, pushing forward both benchmarks and methodologies.
With the proposed datasets, researchers develop various methods to tackle the challenges of AG-ReID.
Wang et al.~\cite{dyu2024dynamic} propose a dynamic token selection transformer to adaptively select informative tokens for cross-view matching, while Hu et al.~\cite{hu2025latex} leverage attribute-based text knowledge to enhance view-invariant representations.
However, these methods primarily focus on extracting view-shared features while discarding view-specific ones.
Different from previous methods, we leverage generative models to use view-specific features for imitating cross-view distributions, thereby enhancing feature robustness.
\begin{figure*}[t]
\centering
\includegraphics[width=0.88\linewidth]{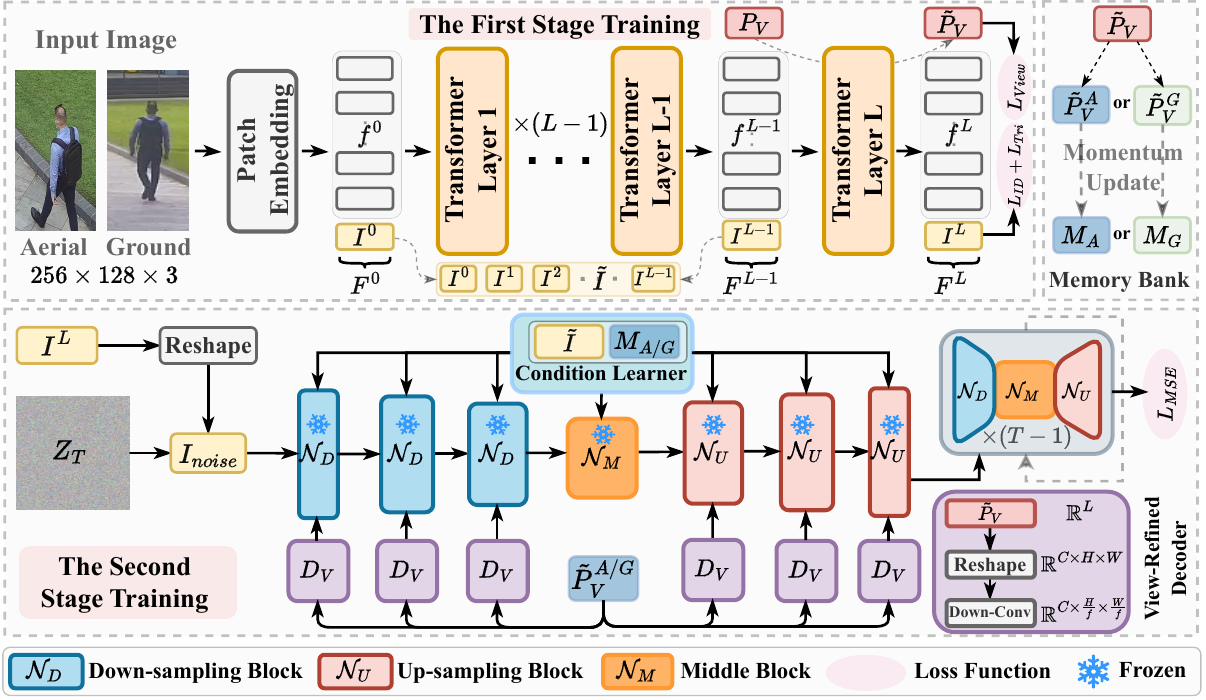}
\vspace{-2mm}
\caption{
Overall framework of the proposed SD-ReID, which follows a two-stage training pipeline with the backbone frozen in Stage~II to preserve the learned identity representations.
In Stage~I, a view-aware Transformer encoder extracts person representations \(I^L\) and instance-level view features \(\tilde{P}_V\), while global view prototypes \(M_A\) and \(M_G\) are maintained in the memory bank as stable view conditions.
In Stage~II, Stable Diffusion is trained to generate view-specific features. The condition learner fuses intermediate representations \(\tilde{I}\) with global view prototypes to guide the generation, while the proposed VRD injects instance-level view features at multiple scales.
During inference, unavailable cross-view features are replaced by global prototypes from the memory bank.
With the proposed two-stage training and modules, SD-ReID effectively enhances the discriminative ability of person representations for AG-ReID.
}
\label{fig:overall}
\vspace{-3mm}
\end{figure*}
\subsection{Diffusion Model}
Recently, diffusion models have emerged as a powerful approach for generating diverse contents, including images, videos and audio.
Their core idea is to iteratively transform simple noise distributions into complex data distributions through successive denoising steps.
In addition to unconditional generative methods~\cite{ho2020denoising,song2020denoising}, numerous techniques have been developed to incorporate additional control signals, enabling more controllable content generation~\cite{dhariwal2021diffusion,ho2022classifier,pandey2022diffusevae}.
However, pixel-level diffusion and denoising are computationally expensive~\cite{saharia2022photorealistic, ramesh2022hierarchical,ho2022cascaded}.
To address this, Latent Diffusion Models (LDMs) compress inputs using a Variational AutoEncoder (VAE) and perform diffusion in the latent space, significantly reducing training and inference costs.
Building on this idea, pre-trained LDMs such as Stable Diffusion~\cite{rombach2022high} integrate a VAE to make diffusion and denoising more efficient.
Beyond generative tasks, recent works employ diffusion models to discriminative tasks~\cite{chen2023diffusiondet, wang2022feature,kim2024pose}, demonstrating strong cross-task generalization capabilities.
Motivated by these successes, in this paper, we leverage diffusion models to directly generate view-specific features.
Furthermore, we propose a novel framework named SD-ReID for AGReID, which integrates both generative and discriminative models for more robust representation learning.
\section{Background of Stable Diffusion}
\label{sec:Preliminaries}
We briefly review the SD model, which leverages the Denoising Diffusion Probabilistic Model (DDPM)~\cite{ho2020denoising} as its training strategy.
Given a real image \(\mathcal{V}_0\), a VAE encoder \(\mathcal{I}(\cdot)\) maps it to a latent vector \(z_0 = \mathcal{I}(\mathcal{V}_0)\), following a distribution \(q(z_0)\).
The forward diffusion gradually adds Gaussian noise \(\epsilon \sim \mathcal{N}(0,\boldsymbol{I})\) over \(T\) timesteps with the following formulation:

\begin{equation}
q(z_{1:T}|z_0) = \prod_{t=1}^{T} q(z_{t}|z_{t-1}),
\end{equation}
\begin{equation}
  q(z_{t}|z_{t-1}) = \mathcal{N} ( z_{t}; \sqrt{1-\beta_{t}}z_{t-1}, \beta_t \boldsymbol{\mathit{I}}),
\end{equation}
where \(\beta_t\) denotes a variance schedule.
For DDPM, the above noising process can be simplified as follows:
\begin{equation}
    z_t = \sqrt{\bar{\alpha}_t} z_0 + \sqrt{1-\bar{\alpha}_t} \epsilon,
\end{equation}
with \(\bar{\alpha}_t\) derived from the fixed schedule.
After \(T\) steps, \(z_T\) becomes an isotropic Gaussian distribution \(\mathcal{N}(0,\boldsymbol{I})\).
Then, the denoising process learns a conditional model to reconstruct \(z_0\) from noisy latent \(z_t\) with the following formulation:
\begin{equation}
    p_\theta(z_{t-1}|z_t,c) = \mathcal{N}( z_{t-1}; \mu_\theta(z_t,c,t), \Sigma_\theta(z_t,c,t) ),
\end{equation}
where \(c\) is the condition information.
Finally, the model minimizes the Mean Squared Error (MSE) loss between the true noise \(\epsilon\) and the predicted noise \(\epsilon_\theta\) as follows:
\begin{equation}
    \mathcal{L}_{MSE} = \mathbb{E}_{z_0,c,\epsilon,t} \left[ \| \epsilon - \epsilon_\theta(z_t,t,c) \|_2^2 \right].
    \label{eq:mse}
\end{equation}

In SD-ReID, we adopt this framework to generate view-specific person representations for AG-ReID.
\section{Proposed Method}
\label{sec:Method}
As shown in \newcref{fig:overall}, SD-ReID follows a two-stage training process.
In the first stage, a ViT-based ReID model extracts person features and control conditions.
In the second stage, with the ReID model fixed, the SD model is trained to generate view-specific features conditioned on identity and view conditions.
Specifically, the view conditions refer to the global view prototypes \(M_A\) and \(M_G\), maintained by a momentum memory bank.
These prototypes guide the diffusion model to generate complementary features for the opposite view, thereby reducing the cross-view distribution gap.
With the two stages, our proposed SD-ReID effectively enhances the discriminative ability of person representations for AG-ReID.
Details are described in the following subsections.
\subsection{The First Stage Training}
\label{subsec:first_stage}
The goal of the first stage is to train a discriminative model that extracts person representations and corresponding control conditions.
As illustrated in \newcref{fig:overall}, we adopt a view-aware visual encoder based on ViT.
Given an input image \(\mathcal{V} \in \mathbb{R}^{H \times W \times 3}\), it is first embedded into class and patch tokens:
\begin{equation}
F^{0} = [I^{0}, f^{0}],
\end{equation}
where $I^{0} \in \mathbb{R}^{C}$ is the class token and $f^{0} \in \mathbb{R}^{N \times C}$ are patch tokens.
$C$ is the embedding dimension.
$N$ is the number of patches.
To incorporate view information, we append a learnable view token $P_V$ to the token sequence before the final Transformer layer $\varOmega_L$ with the following equation:
\begin{equation}
[I^L, f^L, \tilde{P}_V] = \varOmega_L([F^{L-1}, P_V]),
\end{equation}
where \(\tilde{P}_V\) is the instance-level view feature produced by the interaction between \(P_V\) and image features \(F^{L-1}\).
The output class token \(I^L\) serves as the person representation.
Besides, the input to the final layer \(F^{L-1}\) is obtained by stacking the preceding Transformer layers as follows:
\begin{equation}
F^{L-1} = \varOmega_{L-1}(\varOmega_{L-2}(\dots\varOmega_1(F^0)\dots)).
\end{equation}
During training, both the person representation and the instance-level view feature are supervised to ensure discrimination and view-awareness.
However, during inference, the instance-level view feature \(\tilde{P}_V\) is unavailable since the image of the same person with other views is inaccessible in real retrieval scenarios.
To obtain stable view conditions for retrieval, we maintain a momentum-based memory bank that aggregates instance-level view features into global prototypes across the training set.
As shown in the right corner of \newcref{fig:overall}, the global prototypes are updated with the corresponding instance-level view feature with the following equations:
\begin{equation}
M_A \leftarrow \alpha M_A + (1-\alpha)\tilde{P}^{A}_V,
\end{equation}
\begin{equation}
M_G \leftarrow \alpha M_G + (1-\alpha)\tilde{P}^{G}_V,
\end{equation}
where \(\alpha\) is the momentum coefficient.
\(M_A\) and \(M_G\) denote the global view prototypes for aerial and ground views, respectively.
These prototypes provide robust cross-instance view conditions, which are employed to guide the generative process in the second stage as described in \newcref{subsec:second_stage}.
\subsection{The Second Stage Training}
\label{subsec:second_stage}
While the first stage provides discriminative person representations, these features are restricted to the observed view and therefore fail to capture the distribution of other views.
This limitation leads to an incomplete representation, since aerial and ground images of the same person can differ drastically in viewpoint and context.
To bridge this gap, the second stage training aims to imitate the distribution of other view representations by leveraging the SD model.
In this way, the model can generate view-specific features that complement the original representations and enhance cross-view retrieval.

To this end, we perform diffusion in the feature space rather than the pixel space.
This avoids synthesizing high-resolution images and the costly re-encoding through the visual encoder~\cite{rombach2022high}.
It also allows the model to operate directly on identity-discriminative semantics, bypassing irrelevant low-level details such as background texture and illumination.

\begin{figure}[t]
    \centering
        \includegraphics[width=0.84\linewidth]{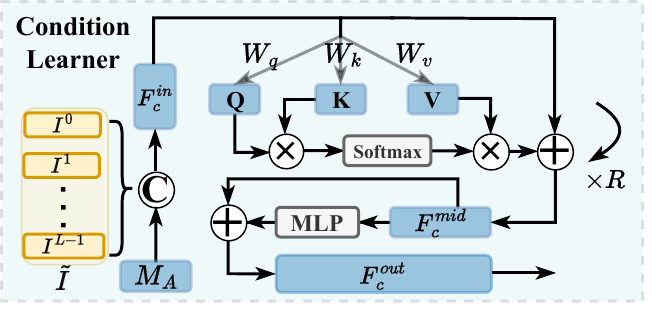}
        \vspace{-2mm}
        \caption{Details of the condition learner based on aerial input.}
        \label{fig:detail}
        \vspace{-4mm}
\end{figure}

However, a key challenge lies in how to effectively condition the diffusion model when explicit descriptors such as text annotations are unavailable, which is typically the case in AG-ReID datasets and real-world scenarios.
To overcome this limitation, we exploit the intermediate class tokens \(I^i\) from each Transformer layer as identity descriptors, since they preserve fine-grained identity cues across different levels.
These descriptors are concatenated to form a comprehensive representation \(\tilde{I}\).
This multi-layer design yields richer identity conditions. Earlier layers capture low-level appearance cues such as color and texture, while deeper layers encode higher-level semantic identity information~\cite{he2021transreid}.
In parallel, global view prototypes maintained by the momentum memory bank are employed as stable view conditions.
By jointly leveraging \(\tilde{I}\) and the global view prototypes, the diffusion model is guided to generate features that are both identity-preserving and view-aware.
To achieve an effective fusion of identity and view information, we introduce a condition learner.
As illustrated in \newcref{fig:detail}, it concatenates the intermediate descriptors with the retrieved global view condition to form the input sequence \(F_{c}^{in}\), which is subsequently refined by \(R\) Transformer layers:
\begin{equation}
    Q = W_q F_{c}^{in}, \quad K = W_k F_{c}^{in}, \quad V = W_v F_{c}^{in},
\end{equation}
\begin{equation}
    F_{c}^{mid} = F_{c}^{in} + \mathcal{S}\!\left(\frac{QK^{T}}{\sqrt{d}}\right)V ,
\end{equation}
\begin{equation}
    F_{c}^{out} = F_{c}^{mid} + \mathcal{M}(F_{c}^{mid}),
\end{equation}
where $W_q$, $W_k$ and $W_v$ are projection matrices.
\(\mathcal{S}\) denotes the Softmax operator and \(\mathcal{M}\) is a multi-layer perceptron (MLP).
This prevents the model from over-relying on any single condition and ensures a stable feature generation under incomplete or noisy cross-view information.
\subsection{View-Refined Decoder}
\label{subsec:VRD}
As discussed in \newcref{subsec:second_stage}, instance-level view features \(\tilde{P}_V\) provide the most accurate guidance for generating other view representations, since they capture fine-grained and image-specific view information.
However, such cross-view features are unavailable during inference (e.g., the ground-view feature is missing when only an aerial input is provided), making it impossible to generate instance-specific features for unseen perspectives.
Even though we construct global-level view prototypes using a memory bank across training set (\newcref{subsec:first_stage}), these features only provide approximate guidance and cannot fully capture instance-specific variations, leading to a distribution gap between training and inference.
To mitigate this issue, we introduce the View-Refined Decoder (VRD) \( D_V \), which is integrated into the down-sampling and up-sampling blocks of the SD model to refine generated features.

\begin{figure}[t]
\centering
\includegraphics[width=0.84\linewidth]{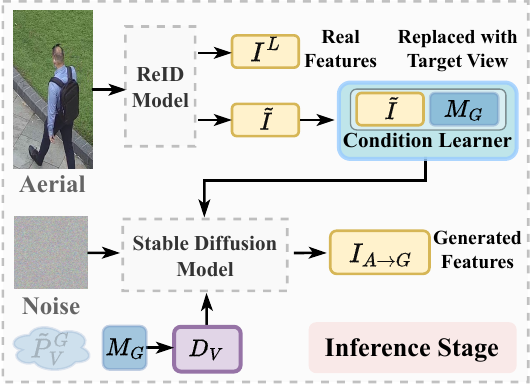}
\caption{Inference process from aerial input to ground view feature generation.}
\label{fig:inference}
\vspace{-2mm}
\end{figure}

Specifically, during training, VRD processes instance-level features \( \tilde{P}_V \), embedding multi-scale view-aware cues that are aligned with feature maps in the SD model.
However, cross-view instance features are unavailable during inference.
As illustrated in \newcref{fig:inference}, taking the inference process from aerial input to ground-view feature generation as an example, VRD replaces the instance-level feature \( \tilde{P}^{G}_V \) with the global prototype \( M_G \) retrieved from the memory bank.
This design allows the model to leverage precise instance-level information during training while remaining feasible at inference, reducing the distribution gap introduced by relying solely on global-level conditions.
Formally, each VRD module is a lightweight feature transformation unit.
It reshapes the input and applies down-convolution operations to produce a feature compatible with the SD model.
By maintaining the compactness of features and minimizing information loss during downsampling, VRD ensures that generated view-specific features remain discriminative and robust for better cross-view retrieval.
\subsection{Optimization}
\label{subsec:opt}
As illustrated in Fig.~\ref{fig:overall}, multiple loss functions are utilized to optimize our framework.
For the ReID model, the label smoothing cross-entropy loss~\cite{szegedy2016rethinking} and triplet loss~\cite{hermans2017defense} are employed to supervise both the discriminative model and the view classifier.
The loss functions can be formulated as:
\begin{equation}
    \mathcal{L}_{ReID} = \mathcal{L}_{ID} + \mathcal{L}_{Tri},
    \label{eq:reidloss}
\end{equation}
\begin{equation}
    \mathcal{L}_{View} =  - \frac{1}{|B|} \sum_{i=1}^{|B|}  v_i \log(\hat{v}_i),
    \label{eq:viewloss}
\end{equation}
where \( B \) is the batch size.
\(v_{i}\) and \(\hat{v}_{i}\) denote the ground truth and corresponding view predictions, respectively.
Thus, the total loss of the first stage can be written as:
\begin{equation}
    \mathcal{L}_{\text{Stage~I}} = \mathcal{L}_{ReID} + \mathcal{L}_{View}.
    \label{eq:stg1loss}
\end{equation}
For the second stage, we convert the learning task to a source-to-source self-reconstruction task.
As described in \newcref{subsec:VRD}, we replace the condition item \( c \) of \newcref{eq:mse} with two condition embeddings, namely \(F_{c}^{out} \) and \(F_{c}^{VRD}\).
Consequently, the total loss of the second stage is formulated as:
\begin{equation}
    \scalebox{0.85}[0.95]{$
    \begin{aligned}
    \mathcal{L}_{\text{Stage~II}} = \mathbb{E}_{z_{0},F_{c}^{out},F_{c}^{VRD},\epsilon,t}[{|| \epsilon - \epsilon_{\theta} (z_t,t,F_{c}^{out},F_{c}^{VRD}) ||}_{2}^{2} ].
    \end{aligned}
    $}
    \label{eq:stg2loss}
\end{equation}
Furthermore, to accelerate convergence, we follow the strategy in \cite{lu2024coarse} by adopting a cubic schedule for the timestep distribution.
Specifically, we first sample \( u \) uniformly from the interval \( [1, T] \), and then set \( t = ( 1 - (\frac{u}{T})^3) \times T \) for better training.
\subsection{Inference}
\label{subsec:inf}
\textbf{Sampling Process.}
In the second stage, the generative model performs inference by sampling from Gaussian noise \( \mathcal{N}(0, \boldsymbol{\mathit{I}}) \).
The cumulative classifier-free guidance~\cite{ho2022classifier} is then applied to reinforce the condition signal.

\textbf{Retrieval Process.}
The complete retrieval process proceeds as follows.
The frozen backbone first extracts visual features \(I^L\) and identity descriptors \(\tilde{I}\).
Global view prototypes \(M_A\)/\(M_G\) are then retrieved from the memory bank to serve as view conditions.
They replace the unavailable instance-level view features \(\tilde{P}_V\) as described in \newcref{subsec:VRD}.
The conditioned SD model generates view-specific features, which are subsequently refined through the VRD.
Finally, the generated and real features are concatenated and L2-normalized for cosine distance computation.
Features are generated for \textit{all} samples, since the view of the matching target is unknown in practice.
\section{Experiments}
\label{sec:Exp}
\begin{table*}[t]
  \centering
  \renewcommand{\arraystretch}{1.22}
  \caption{Comparison with existing methods on CARGO.  %
  The best performance is shown in \textbf{bold} and the second best is \underline{underlined}.
  }
  \vspace{-2mm}
  \resizebox{1.62\columnwidth}{!}{
  \begin{tabular}{c c c c c c c c c c c}
    \toprule
    \multirow{2}{*}{Method}
    & \multicolumn{2}{c}{A\(\leftrightarrow\)G}
    & \multicolumn{2}{c}{ALL}
    & \multicolumn{2}{c}{G\(\leftrightarrow\)G}
    & \multicolumn{2}{c}{A\(\leftrightarrow\)A}
    & \multicolumn{2}{c}{Average} \\
    \cline{2-11}
    ~ & Rank-1 & mAP
       & Rank-1 & mAP
       & Rank-1 & mAP
       & Rank-1 & mAP
       & Rank-1 & mAP \\
    \midrule
    SBS~\cite{he2023fastreid} & 31.25 & 29.00 & 50.32 & 43.09 & 72.31 & 62.99 & 67.50 & 49.73 & 55.35 & 46.20 \\
    PCB~\cite{sun2019learning} & 34.40 & 30.40 & 51.00 & 44.50 & 74.10 & 67.60 & 55.00 & 44.60 & 53.63 & 46.78 \\
    BoT~\cite{luo2019bag} & 36.25 & 32.56 & 54.81 & 46.49 & 77.68 & 66.47 & 65.00 & 49.79 & 58.44 & 48.83 \\
    MGN~\cite{wang2018learning} & 31.87 & 33.47 & 54.81 & 49.08 & \textbf{83.93} & 71.05 & 65.00 & 52.96 & 58.90 & 51.64 \\
    VV~\cite{kumar2020strong} & 31.25 & 29.00 & 45.83 & 38.84 & 72.31 & 62.99 & 67.50 & 49.73 & 54.22 & 45.14 \\
    AGW~\cite{ye2021deep} & 43.57 & 40.90 & 60.26 & 53.44 & 81.25 & 71.66 & 67.50 & 56.48 & 63.15 & 55.62 \\
    ViT~\cite{dosovitskiy2020image} & 43.13 & 40.11 & 61.54 & 53.54 & 82.14 & 71.34 & 80.00 & 64.47 & 66.70 & 57.37 \\
    VDT~\cite{zhang2024view} & 45.00 & 42.08 & 60.58 & 54.61 & 76.79 & 71.97 & 82.50 & 64.67 & 66.22 & 58.33 \\
    DTST~\cite{dyu2024dynamic} & \underline{50.63} & 43.39  & 64.42 & 55.73  & 78.57 & 72.40   & 80.00 & 63.31  & 68.41 & 58.71 \\
    SeCap~\cite{wang2025secap} & 48.75 & \underline{46.37} & \underline{64.72} & \underline{56.89} & \underline{82.54} & \textbf{75.24} & \underline{82.50} & \underline{66.90} & \underline{69.63} & \underline{61.35} \\
    \midrule
    \rowcolor[gray]{0.9}
    \textbf{SD-ReID} & \textbf{53.12} & \textbf{46.44} & \textbf{65.06} & \textbf{57.47} & 81.25 & \underline{74.08} & \textbf{82.50} & \textbf{67.70} & \textbf{70.48} & \textbf{61.42} \\
    \bottomrule
  \end{tabular}
  }
  \vspace{-2mm}
  \label{tab:datacargo}
\end{table*}
\begin{table}[t]
  \centering
  \renewcommand{\arraystretch}{1.22}
  \caption{Comparison with existing methods on AG-ReID.v1.
  }
  \vspace{-2mm}
  \resizebox{0.9\columnwidth}{!}{
  \begin{tabular}{c  c c  c c  c c}
      \toprule
      \multirow{2}{*}{Method} & \multicolumn{2}{c}{A→G} & \multicolumn{2}{c}{G→A} & \multicolumn{2}{c}{Average} \\
      \cmidrule(r){2-3}\cmidrule(r){4-5}\cmidrule(r){6-7}
       ~ & Rank-1 & mAP & Rank-1 & mAP & Rank-1 & mAP \\
      \midrule
      OSNet\cite{zhou2021learning}     & 72.59 & 58.32 & 74.22 & 60.99 & 73.41 & 59.66 \\
      BoT\cite{luo2019bag}             & 70.01 & 55.47 & 71.20 & 58.83 & 70.61 & 57.15 \\
      SBS\cite{he2023fastreid}         & 73.54 & 59.77 & 73.70 & 62.27 & 73.62 & 61.02 \\
      VV\cite{kumar2020strong}         & 77.22 & 67.23 & 79.73 & 69.83 & 78.48 & 68.53 \\
      ViT\cite{dosovitskiy2020image}   & 81.28 & 72.38 & 82.64 & 73.35 & 81.96 & 72.87 \\
      TransReID\cite{he2021transreid}& 81.80 & 73.10 & 83.40 & 74.60 & 82.60 & 73.85 \\
      FusionReID\cite{wang2025unity}   & 80.40 & 71.40 & 82.40 & 74.20 & 81.40 & 72.80 \\
      CLIP-ReID\cite{li2023clip}       & 79.44 & 70.55 & 84.20 & 73.05 & 81.82 & 71.80 \\
      PCL-CLIP\cite{li2023prototypical}& 82.16 & 73.11 & \textbf{86.90} & 76.28 & \underline{84.53} & 74.70 \\
      Explain\cite{nguyen2023aerial}   & 81.47 & 72.61 & 82.85 & 73.39 & 82.16 & 73.00 \\
      VDT \cite{zhang2024view}         & 83.00 & 74.06 & 84.62 & \underline{76.28} & 83.81 & 75.17 \\
      DTST\cite{dyu2024dynamic}        & \underline{83.48} & \underline{74.51} & 84.72 & 76.05 & 84.10 & \underline{75.28} \\
      SeCap~\cite{wang2025secap}       & 81.13 & 72.51 & 84.10 & 75.45 & 82.62 & 73.98 \\
      \midrule
      \rowcolor[gray]{0.9}
      \textbf{SD-ReID}                 & \textbf{85.16} & \textbf{75.40} & \underline{85.97} & \textbf{77.02} & \textbf{85.57} & \textbf{76.21} \\
      \bottomrule
  \end{tabular}
  }
  \label{tab:datav1_avg}
  \vspace{-3mm}
\end{table}
\begin{table*}[t]
    \centering
    \renewcommand{\arraystretch}{1.22}
    \caption{Performance comparison with existing methods on AG-ReID.v2.}
    \vspace{-2mm}
    \resizebox{1.60\columnwidth}{!}{
    \begin{tabular}{c c  c c c c c c c  c c}
      \toprule
      \multirow{2}{*}{Method}  & \multicolumn{2}{c}{A→C} & \multicolumn{2}{c}{A→W} & \multicolumn{2}{c}{C→A} & \multicolumn{2}{c}{W→A} & \multicolumn{2}{c}{Average} \\
      \cmidrule(r){2-3} \cmidrule(r){4-5} \cmidrule(r){6-7} \cmidrule(r){8-9} \cmidrule(r){10-11}
      ~ & Rank-1 & mAP & Rank-1 & mAP & Rank-1 & mAP & Rank-1 & mAP & Rank-1 & mAP \\
      \midrule
      Swin\cite{liu2021swin} & 68.76 & 57.66 & 68.49 & 56.15 & 68.80 & 57.70 & 64.40 & 53.90 & 67.61 & 56.35 \\
      HRNet-18\cite{wang2020deep} & 75.21 & 65.07 & 76.26 & 66.17 & 76.25 & 66.16 & 76.25 & 66.17 & 75.99 & 65.89 \\
      SwinV2\cite{liu2022swin} & 76.44 & 66.09 & 80.08 & 69.09 & 77.11 & 62.14 & 74.53 & 65.61 & 77.04 & 65.73 \\
      MGN(R50)\cite{wang2018learning} & 82.09 & 70.17 & 88.14 & 78.66 & 84.21 & 72.41 & 84.06 & 73.73 & 84.63 & 73.74 \\
      BoT(R50)\cite{luo2019bag} & 80.73 & 71.49 & 86.06 & 75.98 & 79.46 & 69.67 & 82.69 & 72.41 & 82.24 & 72.39 \\
      BoT(R50)+Attributes & 81.43 & 72.19 & 86.66 & 76.68 & 80.15 & 70.37 & 83.29 & 73.11 & 82.88 & 73.09 \\
      SBS(R50)\cite{he2023fastreid} & 81.96 & 72.04 & 88.14 & 78.94 & 84.10 & 73.89 & 84.66 & 75.01 & 84.72 & 74.97 \\
      SBS(R50)+Attributes & 82.56 & 72.74 & 88.74 & 79.64 & 84.80 & 74.59 & 85.26 & 75.71 & 85.34 & 75.67 \\
      ViT\cite{dosovitskiy2020image} & 85.40 & 77.03 & 89.77 & 80.48 & 84.65 & 75.90 & 84.27 & 76.59 & 86.02 & 77.50 \\
      PCL-CLIP\cite{li2023prototypical} & 79.80 & 72.20 & 87.14 & 77.70 & 81.12 & 72.40 & 84.19 & 73.89 & 83.06 & 74.05 \\
      Explain\cite{nguyen2023ag} & \textbf{87.70} & 79.00 & \textbf{93.67} & \underline{83.14} & \textbf{87.35} & \underline{78.24} & \textbf{87.73} & \underline{79.08} & \textbf{89.11} & 79.87 \\
      VDT\cite{zhang2024view} & 86.46 & 79.13 & 90.00 & 82.21 & 86.14 & 78.12 & 85.26 & 78.52 & 86.97 & 79.50 \\
      SeCap~\cite{wang2025secap} & 86.88 & \underline{80.02} & 90.06 & 82.89 & 85.97 & 78.15 & 85.17 & 78.54 & 87.02 & \underline{79.90} \\
      \midrule
      \rowcolor[gray]{0.9}
      \textbf{SD-ReID} & \underline{87.04} & \textbf{80.61} & \underline{90.86} & \textbf{84.06} & \underline{86.74} & \textbf{79.24} & \underline{86.79} & \textbf{80.12} & \underline{87.86} & \textbf{81.01} \\
      \bottomrule
    \end{tabular}
    }
    \vspace{-2mm}
    \label{tab:datav2_avg}
\end{table*}
\begin{table*}[t]
  \centering
  \renewcommand{\arraystretch}{1.22}
  \caption{Performance comparison on LAGPeR and G2APS-ReID datasets.
  CLIP-ReID* indicates using OLP and SIE in CLIP-ReID. MIP$^{\dagger}$ represents the re-implementation for AGReID.
  }
  \vspace{-2mm}
  \resizebox{2\columnwidth}{!}{
  \begin{tabular}{c c c c c c c c c c c c c c c}
    \toprule
    \multirow{2}{*}{Method}
    & \multicolumn{8}{c}{LAGPeR}
    & \multicolumn{6}{c}{G2APS-ReID} \\
    \cline{2-15}
    ~
      & \multicolumn{2}{c}{A$\rightarrow$G}
      & \multicolumn{2}{c}{G$\rightarrow$A}
      & \multicolumn{2}{c}{G$\rightarrow$A+G}
      & \multicolumn{2}{c}{Average}
      & \multicolumn{2}{c}{A$\rightarrow$G}
      & \multicolumn{2}{c}{G$\rightarrow$A}
      & \multicolumn{2}{c}{Average} \\
    \cline{2-15}
    ~
      & Rank-1 & mAP
      & Rank-1 & mAP
      & Rank-1 & mAP
      & Rank-1 & mAP
      & Rank-1 & mAP
      & Rank-1 & mAP
      & Rank-1 & mAP \\
    \midrule
    ViT~\cite{dosovitskiy2020image}
      & 38.67 & 27.25 & 32.04 & 30.69 & 18.88 & 15.31
      & 29.86 & 24.42
      & 69.38 & 52.17 & 67.16 & 52.22
      & 68.27 & 52.20 \\
    TransReID~\cite{he2021transreid}
      & 38.80 & 28.80 & 33.00 & 32.10 & \textbf{22.90} & \underline{18.80}
      & 31.57 & 26.57
      & 67.10 & 53.10 & 68.52 & 54.19
      & 67.81 & 53.65 \\
    CLIP-ReID~\cite{li2023clip}
      & 24.40 & 17.60 & 21.30 & 20.80 & 12.30 & 10.20
      & 19.33 & 16.20
      & 58.30 & 42.20 & 56.41 & 41.92
      & 57.36 & 42.06 \\
    CLIP-ReID*~\cite{li2023clip}
      & 23.10 & 17.50 & 20.00 & 20.30 & 9.00 & 8.40
      & 17.37 & 15.40
      & 59.60 & 42.70 & 56.39 & 42.52
      & 58.00 & 42.61 \\
    MIP$^{\dagger}$~\cite{wu2024enhancing}
      & 39.30 & 29.30 & 33.90 & 32.60 & 21.00 & 17.30
      & 31.40 & 26.40
      & \underline{73.00} & \textbf{57.40} & \textbf{70.22} & \textbf{57.06}
      & \underline{71.61} & \textbf{57.23} \\
    Explain~\cite{nguyen2023aerial}
      & \underline{40.48} & 28.89 & 32.96 & 31.91 & 22.03 & 17.89
      & 31.82 & 26.23
      & 70.75 & 52.87 & 68.70 & 53.39
      & 69.73 & 53.13 \\
    VDT~\cite{zhang2024view}
      & 39.59 & 28.82 & 34.13 & 32.10 & 22.78 & 18.24
      & 32.17 & 26.39
      & 71.45 & 54.13 & 66.84 & 52.71
      & 69.15 & 53.42 \\
    SeCap~\cite{wang2025secap}
      & \textbf{40.51} & \underline{29.59} & \underline{33.95} & \underline{32.60} & 22.23 & 17.97
      & \underline{32.23} & \underline{26.72}
      & 72.31 & 56.51 & 68.15 & 55.12
      & 70.23 & 55.82 \\
    \midrule
    \rowcolor[gray]{0.9}
    \textbf{SD-ReID}
      & 40.15 & \textbf{29.72} & \textbf{34.47} & \textbf{32.86} & \underline{22.85} & \textbf{18.88}
      & \textbf{32.49} & \textbf{27.15}
      & \textbf{73.73} & \underline{56.89} & \underline{69.91} & \underline{56.38}
      & \textbf{71.82} & \underline{56.64} \\
    \bottomrule
  \end{tabular}
  }
  \vspace{-2mm}
  \label{tab:lagper_g2aps_avg}
\end{table*}
\begin{table}[t]
  \centering
  \caption{Performance comparison of SD-ReID's stages and baselines.}
  \vspace{-2mm}
  \resizebox{1\columnwidth}{!}{
  \renewcommand{\arraystretch}{1.22}
  \begin{tabular}{c c  c c c  c c c }
      \toprule
      \multirow{2}{*}{Methods} & \multicolumn{3}{c}{A→G} & \multicolumn{3}{c}{G→A} \\
      \cmidrule(r){2-4}\cmidrule(r){5-7}
      ~ & Rank-1 & mAP & mINP & Rank-1 & mAP & mINP \\
      \midrule
      ViT\cite{dosovitskiy2020image}           & 81.28 & 72.38 & $-$ & 82.64 & 73.35 & $-$ \\

      Explain\cite{nguyen2023aerial}       & 81.47 & 72.61 & $-$ & 82.85 & 73.39 & $-$ \\

      VDT \cite{zhang2024view}          & 83.00 & 74.06 & 50.31 & 84.62 & 76.28 & 49.51 \\
      \midrule
      \rowcolor[gray]{0.9}
      SD-ReID(Stage~I) & 84.60 & 74.86 & 50.32 & 85.65 & 76.85 & 50.05 \\
      \rowcolor[gray]{0.9}
      SD-ReID(Stage~II) & \textbf{85.16} & \textbf{75.40} & \textbf{51.35} & \textbf{85.97} & \textbf{77.02} & \textbf{50.42} \\
      \bottomrule
      \textbf{$\uparrow$} & \textbf{0.56} & \textbf{0.54} & \textbf{1.03} & \textbf{0.32} & \textbf{0.17} & \textbf{0.37} \\
      \textbf{$\Uparrow$} & \textbf{2.16} & \textbf{1.34} & \textbf{1.04} & \textbf{1.35} & \textbf{0.74} & \textbf{0.91} \\
      \bottomrule
      \end{tabular}
  }
  \label{tab:ablav1}
  \vspace{-2mm}
\end{table}
\begin{table}[t]
  \centering
  \caption{Performance comparison with employing real features and generated features.
  Gen($\cdot$) denotes the generated features under corresponding view.
  }
  \renewcommand{\arraystretch}{1.22}
  \resizebox{1\columnwidth}{!}{
  \begin{tabular}{c  c c c  c c c}
      \hline
      \multirow{2}{*}{Features} & \multicolumn{3}{c}{A→G} & \multicolumn{3}{c}{G→A} \\
      \cmidrule(r){2-4}  \cmidrule(r){5-7}
       ~ & Rank-1 & mAP & mINP & Rank-1 & mAP & mINP \\
      \hline
      Real & 84.60 & 74.86 & 50.32 & 85.65 & 76.85 & 50.05 \\
      Gen(A) & 84.51 & 74.40 & 49.70 & 83.99 & 76.00 & 49.66 \\
      Gen(G) & 84.41 & 74.43 & 50.37 & 84.20 & 75.99 & 50.17 \\
      Gen(AG) & 84.88 & 74.96 & 50.69 & 84.62 & 76.25 & 49.88 \\
      Real + Gen(AG) & \textbf{85.16} & \textbf{75.40} & \textbf{51.35} & \textbf{85.97} & \textbf{77.02} & \textbf{50.42} \\
      \hline
  \end{tabular}
  }
  \label{tab:datagen}
    \vspace{-2mm}
\end{table}
\subsection{Datasets and Evaluation Protocols}
\textbf{Datasets.}
We evaluate our methods on five AG-ReID benchmarks, including one synthetic dataset (CARGO~\cite{zhang2024view}) and four real-world datasets (AG-ReID.v1~\cite{nguyen2023aerial}, AG-ReID.v2~\cite{nguyen2024ag}, LAGPeR~\cite{wang2025secap} and G2APS-ReID~\cite{wang2025secap}).
CARGO contains 108,563 images of 5,000 identities collected by eight ground cameras and five aerial cameras.
AG-ReID.v1 contains 21,983 images of 388 identities collected from one aerial camera and one ground camera, with aerial views captured at altitudes between 15 and 45 meters.
AG-ReID.v2 extends AG-ReID.v1 by introducing additional viewpoints and more identities.
LAGPeR consists of 63,841 images of 4,231 identities collected from fourteen ground cameras and seven aerial cameras, with aerial viewpoints at altitudes between 20 and 60 meters.
G2APS-ReID is reconstructed from the person search dataset G2APS~\cite{zhang2023ground}, and contains 200,864 images of 2,788 identities captured by one aerial camera and one ground camera.

\textbf{Evaluation Protocols.}
Following common practices in AG-ReID, we adopt Cumulative Matching Characteristic (CMC) at Rank-1~\cite{moon2001computational}, mean Average Precision (mAP)~\cite{zheng2015scalable}, and mean Inverse Negative Penalty (mINP)~\cite{ye2021deep} as evaluation metrics.
For the CARGO dataset, we use four protocols: ``ALL'' for comprehensive evaluation, ``G\(\leftrightarrow\)G'' for ground-to-ground matching, ``A\(\leftrightarrow\)A'' for aerial-to-aerial matching, and ``A\(\leftrightarrow\)G'' for aerial-to-ground matching.
AG-ReID.v1 and G2APS-ReID are evaluated with two cross-view settings: ``A→G'' and ``G→A''.
LAGPeR is evaluated with three settings: ``A→G'', ``G→A'' and ``G→A+G''.
AG-ReID.v2 extends AG-ReID.v1 by introducing two additional viewpoints.
The Wearable view (W) represents a lower viewpoint close to the human eye level.
The CCTV view (C) has a height similar to the ground cameras (G) used in other datasets. Based on these views, AG-ReID.v2 defines four protocols: ``A→C'', ``A→W'', ``C→A'' and ``W→A''.
In all cases, ``A'' denotes aerial views, ``G'' denotes ground views, ``C'' denotes CCTV views and ``W'' denotes wearable views.
The arrow indicates the retrieval direction from query to gallery, while the double arrow indicates a bidirectional retrieval protocol.
\subsection{Implementation Details}
Our SD-ReID is trained on one NVIDIA A100 GPU and implemented with the PyTorch toolbox and HuggingFace Diffusers~\cite{von2022diffusers}.
In the first stage, we adopt a pre-trained ViT-B/16 as the backbone and resize input images to 256\(\times\)128\(\times\)3.
Data augmentation includes random horizontal flipping, padding and random erasing~\cite{zhong2020random}.
The second stage is built upon Stable Diffusion v1.5~\cite{rombach2022high}, where we follow the common practice and apply random horizontal flipping as data augmentation.
For both training stages, the batch size is set to 128, consisting of 32 identities with 4 instances per identity.
The ReID model is optimized using SGD~\cite{bottou2010large} with a base learning rate of 0.008 for 120 epochs, while the generative model is optimized using Adam~\cite{kingma2014adam} with a base learning rate of 0.0001 for 120 epochs.
The momentum of the memory bank is set to 0.8 as default.
\subsection{Comparison with State-of-the-art Methods}
We evaluate SD-ReID on five AG-ReID benchmarks, including CARGO (\newcref{tab:datacargo}), AG-ReID.v1 (\newcref{tab:datav1_avg}), AG-ReID.v2 (\newcref{tab:datav2_avg}), LAGPeR and G2APS-ReID (\newcref{tab:lagper_g2aps_avg}).
Overall, SD-ReID consistently surpasses other state-of-the-art methods, demonstrating the effectiveness of generating view-specific features for robust cross-view person retrieval.
On the CARGO dataset (\newcref{tab:datacargo}), SD-ReID achieves 53.12\% Rank-1 and 46.44\% mAP under the A\(\leftrightarrow\)G protocol.
Compared with the previous best method, it improves Rank-1 by 2.49\% and mAP by 0.07\%. Across other protocols, SD-ReID either ranks first or second, resulting in an overall average of 70.48\% Rank-1 and 61.42\% mAP.
These results indicate that SD-ReID consistently outperforms competitive baselines like SeCap and DTST, particularly in challenging cross-view scenarios.
On AG-ReID.v1 (\newcref{tab:datav1_avg}), SD-ReID achieves 85.16\% Rank-1 and 75.40\% mAP for A→G, and 85.97\% Rank-1 with 77.02\% mAP for G→A.
Its average performance of 85.57\% Rank-1 and 76.21\% mAP exceeds that of all prior methods, including strong Transformer-based models such as PCL-CLIP and VDT.
These gains confirm that the generated view-specific features effectively enhance cross-view matching, even on smaller datasets.
On AG-ReID.v2 (\newcref{tab:datav2_avg}), which introduces wearable and CCTV viewpoints, SD-ReID achieves the highest mAP across all four protocols and highly competitive Rank-1 scores, with an overall average of 81.01\% mAP and 87.86\% Rank-1.
It consistently outperforms previous methods including Explain and SeCap, demonstrating strong generalization to heterogeneous viewpoints, particularly low-elevation wearable cameras.
On LAGPeR and G2APS-ReID (\newcref{tab:lagper_g2aps_avg}), SD-ReID maintains leading performance.
On LAGPeR, it achieves 32.49\% Rank-1 and 27.15\% mAP, surpassing SeCap and VDT. On G2APS-ReID, it reaches 71.82\% Rank-1 and 56.64\% mAP, showing competitive advantages over other strong baselines.
These results highlight SD-ReID’s effectiveness across datasets of varying scales and complexities.
In summary, SD-ReID improves performance across five benchmarks by generating view-specific features, confirming its robustness under diverse cross-view scenarios.
\subsection{Ablation Studies and Analysis}
In this section, we conduct extensive ablation experiments on AG-ReID.v1 to validate the effectiveness of our proposed modules.
Specifically, our baseline model represents the first stage of SD-ReID, which is a ViT-based AG-ReID model trained with the contrastive loss and ID loss.

\textbf{Effect of the Second Stage Training.}
As shown in \newcref{tab:ablav1}, the first-stage model already achieves competitive performance.
When incorporating generative models in the second stage, further improvements are observed, with Rank-1 increasing by 0.56\% and mINP by 1.03\% under the A\(\rightarrow\)G setting.
This demonstrates that generated features remain beneficial even on small-scale datasets.
These consistent improvements across different metrics clearly validate the effectiveness of generated features in enhancing AG-ReID performance.

\textbf{Effect of Training Strategies.}
\newcref{tab:backbone} compares different training strategies in the second stage.
Jointly fine-tuning the backbone reduces average Rank-1 by 1.63\%.
Simultaneous updates to the input features, generation targets, and memory-bank prototypes destabilize the denoising objective.
In contrast, freezing the backbone anchors a fixed feature space, allowing the diffusion model to converge more stably.
We therefore adopt the frozen-backbone strategy as default.

\begin{table}[t]
  \centering
  \caption{Effect of different backbone training strategies on the second stage performance.}
  \vspace{-2mm}
  \renewcommand{\arraystretch}{1.22}
  \resizebox{0.9\columnwidth}{!}{
  \begin{tabular}{c  c c  c c  c c}
      \toprule
      \multirow{2}{*}{Strategy} & \multicolumn{2}{c}{A$\rightarrow$G} & \multicolumn{2}{c}{G$\rightarrow$A} & \multicolumn{2}{c}{Average} \\
      \cmidrule(r){2-3}\cmidrule(r){4-5}\cmidrule(r){6-7}
      ~ & Rank-1 & mAP & Rank-1 & mAP & Rank-1 & mAP \\
      \midrule
      Joint Fine-tuning & 83.45 & 73.80 & 84.42 & 75.48 & 83.94 & 74.64 \\
      \textbf{Frozen (Ours)} & \textbf{85.16} & \textbf{75.40} & \textbf{85.97} & \textbf{77.02} & \textbf{85.57} & \textbf{76.21} \\
      \bottomrule
  \end{tabular}
  }
  \label{tab:backbone}
  \vspace{-2mm}
\end{table}
\begin{table}[t]
  \centering
\caption{
Performance comparison using person representations of different qualities as generation targets.
* indicates SD-ReID results with low-quality targets.
}
  \renewcommand{\arraystretch}{1.22}
  \resizebox{0.9\columnwidth}{!}{
  \begin{tabular}{c  c c c  c c c}
      \toprule
      \multirow{2}{*}{Method} & \multicolumn{3}{c}{A→G} & \multicolumn{3}{c}{G→A} \\
      \cmidrule(r){2-4}  \cmidrule(r){5-7}
       ~ & Rank-1 & mAP & mINP & Rank-1 & mAP & mINP \\
      \midrule
      Stage~I & 84.60 & 74.86 & 50.32 & 85.65 & 76.85 & 50.05 \\
      Stage~II & 85.16 & 75.40 & 51.35 & 85.97 & 77.02 & 50.42 \\
      \midrule
      Stage~I* & 10.14 & 5.09 & 0.97 & 8.94 & 5.21 & 1.29 \\
      Stage~II* & 8.92 & 4.41 & 0.78 & 8.84 & 4.88 & 1.23 \\
      \bottomrule
  \end{tabular}
  }
  \label{suptab:target}
\end{table}
\begin{figure}[t]
  \hfill
  \centering
      \includegraphics[width=1\linewidth]{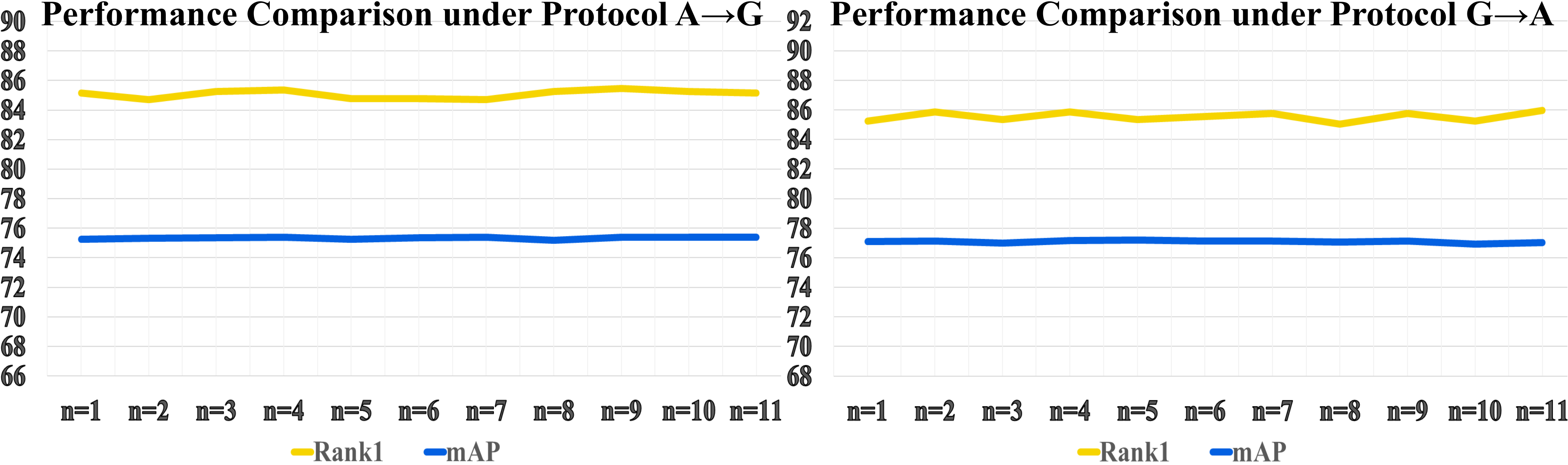}
  \caption{Performance comparison with different numbers of identity conditions under both A→G and G→A protocols.}
  \label{supfig:IDNum}
\end{figure}

\textbf{Effect of Generated Features.}
As shown in \newcref{tab:datagen}, retrieval using only generated features achieves performance highly comparable to real features, with most gaps within 0.5\% and some metrics nearly identical.
This confirms that the generated features are of sufficient quality for independent retrieval.
Moreover, comparable results across different viewpoints indicate that our model introduces no bias toward specific conditions.
Combining features from multiple generated viewpoints further improves performance, in some cases even surpassing real features.
Finally, fusing real and generated features from all viewpoints achieves the best results, reaching 85.16\% Rank-1 and 75.40\% mAP under the A→G setting.
These results confirm the importance of incorporating generated features from all views to mitigate intra-view bias.

\textbf{Effect of Target Representation Quality.}
\newcref{suptab:target} analyzes the impact of using person representations of different quality levels as generation targets in the second stage.
These representations are extracted from the first stage, where low-quality versions are obtained by removing the ReID loss in \newcref{eq:reidloss}.
Results show that low-quality targets degrade the generated features and consequently impair performance.

\textbf{Effect of the Number of Identity Conditions.}
\newcref{supfig:IDNum} shows the impact of varying the number of ID conditions extracted from the first-stage model (starting at the 11-th layer) on performance.
Overall, both evaluation protocols exhibit only minor fluctuations as the number of ID conditions increases, indicating stable performance.
Based on these observations, we adopt all 11 layers of representations as ID conditions.
\begin{figure}[t]
  \centering
     \includegraphics[width=0.89\linewidth]{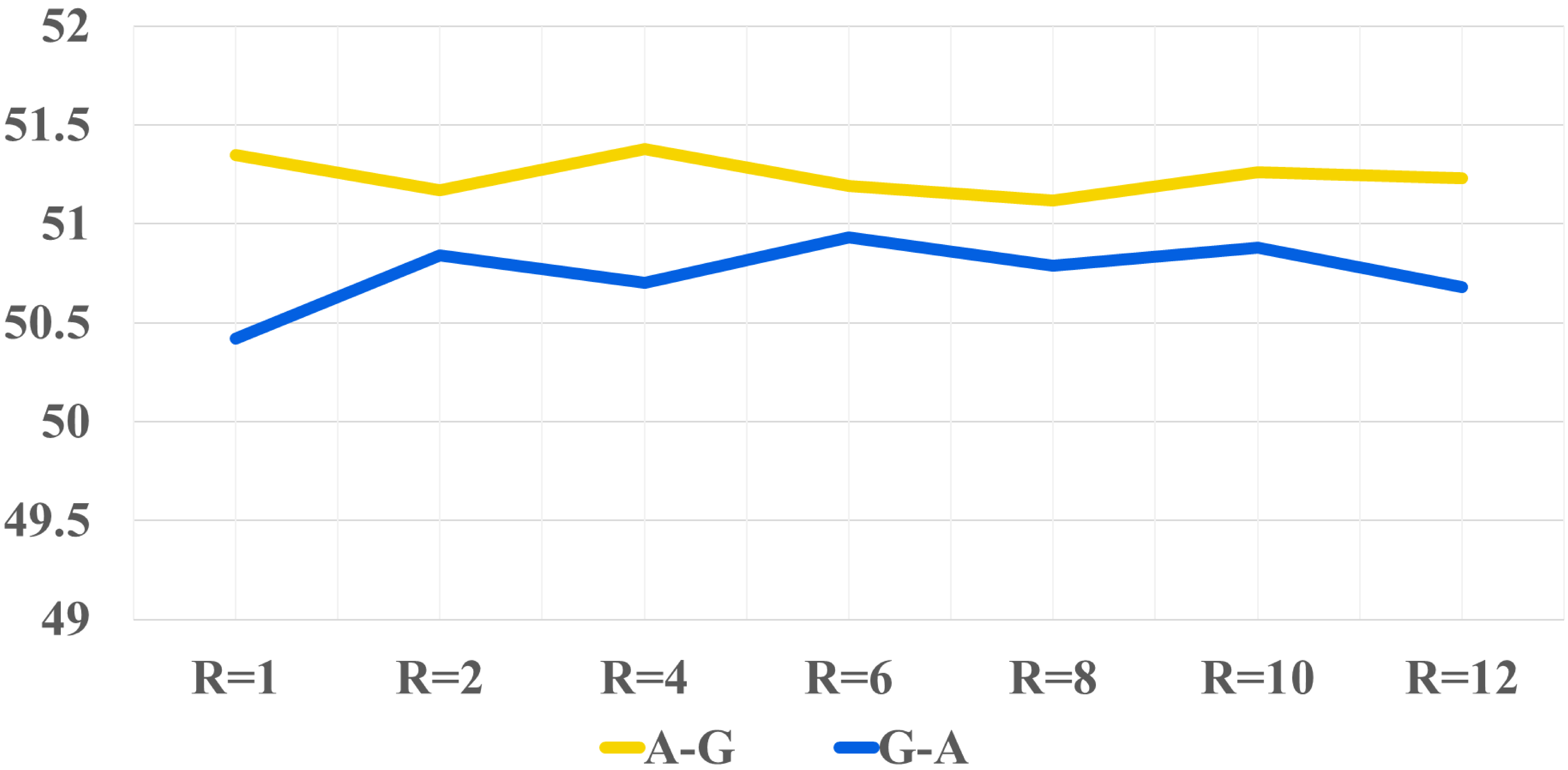}
\caption{mINP comparison with different layer numbers of the condition learner under both A→G and G→A protocols.}
  \label{supfig:layers}
\end{figure}
\begin{table}[t]
\centering
\caption{
Performance comparison using condition features of different qualities.
* indicates results obtained without the view classification loss.
}
  \renewcommand{\arraystretch}{1.22}
  \resizebox{0.9\columnwidth}{!}{
  \begin{tabular}{c  c c c  c c c}
      \toprule
      \multirow{2}{*}{Method} & \multicolumn{3}{c}{A→G} & \multicolumn{3}{c}{G→A} \\
      \cmidrule(r){2-4}  \cmidrule(r){5-7}
       ~ & Rank-1 & mAP & mINP & Rank-1 & mAP & mINP \\
      \midrule
      SD-ReID  & 85.16 & 75.40 & 51.35 & 85.97 & 77.02 & 50.42 \\
      SD-ReID* & 83.76 & 74.67 & 51.26 & 84.62 & 76.76 & 50.70 \\
      \bottomrule
  \end{tabular}
  }
  \label{suptab:ViewCondition}
\end{table}
\begin{table}[t]
  \centering
  \caption{Effect of batch size on memory bank prototype estimation.}
  \vspace{-2mm}
  \renewcommand{\arraystretch}{1.22}
  \resizebox{0.9\columnwidth}{!}{
  \begin{tabular}{c  c c  c c  c c}
      \toprule
      \multirow{2}{*}{Batch Size} & \multicolumn{2}{c}{A$\rightarrow$G} & \multicolumn{2}{c}{G$\rightarrow$A} & \multicolumn{2}{c}{Average} \\
      \cmidrule(r){2-3}\cmidrule(r){4-5}\cmidrule(r){6-7}
      ~ & Rank-1 & mAP & Rank-1 & mAP & Rank-1 & mAP \\
      \midrule
      16 & 83.12 & 73.48 & 84.24 & 75.16 & 83.68 & 74.32 \\
      32 & 84.25 & 74.60 & 85.09 & 76.12 & 84.67 & 75.36 \\
      64 & 85.10 & 75.33 & 85.92 & 76.96 & 85.51 & 76.15 \\
      \textbf{128 (Ours)} & \textbf{85.16} & \textbf{75.40} & \textbf{85.97} & \textbf{77.02} & \textbf{85.57} & \textbf{76.21} \\
      256 & 85.08 & 75.30 & 85.88 & 76.92 & 85.48 & 76.11 \\
      \bottomrule
  \end{tabular}
  }
  \label{tab:batch}
  \vspace{-2mm}
\end{table}

\textbf{Effect of Different Condition Learner Layers.}
We evaluate the impact of varying the number of layers \( R \) in the Condition Learner, as shown in \newcref{supfig:layers}.
Across both protocols, mINP remains stable with fluctuations within 1\%, even when \( R \) varies from 1 to 12.
This demonstrates the robustness of our method to this hyperparameter.
We therefore set \( R = 1 \) as default to balance efficiency and effectiveness in experiments.

\textbf{Effect of View Conditions.}
\newcref{suptab:ViewCondition} shows the effect of view conditions.
By removing the viewpoint classification loss in the first stage, explicit supervision is lost, which degrades the quality of viewpoint features.
Consequently, the generated features are also affected, leading to reduced performance.
These results highlight the critical role of view conditions in the overall effectiveness of our proposed SD-ReID framework.

\textbf{Effect of the Momentum Memory Bank.}
We evaluate the effect of our momentum memory bank from three perspectives, namely batch size, momentum coefficient, and view imbalance.
As shown in \newcref{tab:batch}, reducing the batch size from 128 to 16 lowers average Rank-1 by 1.89\%.
This is because fewer identities per batch degrade both the ReID training signal and the prototype estimation accuracy.
Performance largely stabilizes beyond batch size 64.
\newcref{tab:alpha} further shows that performance is robust for \(\alpha \in [0.8, 0.9]\).
Too small an \(\alpha\) causes rapid prototype drift, while too large a value yields overly conservative updates.
We also simulate view imbalance by keeping all aerial-view data fixed and progressively reducing the ground-view data.
As shown in \newcref{tab:imbalance}, even under a severe 1:0.25 ratio, average Rank-1 drops by only 0.63\%, confirming robustness to view-distribution shifts.

\begin{table}[t]
  \centering
  \caption{Effect of the memory bank momentum coefficient \(\alpha\).}
  \vspace{-2mm}
  \renewcommand{\arraystretch}{1.22}
  \resizebox{0.9\columnwidth}{!}{
  \begin{tabular}{c  c c  c c  c c}
      \toprule
      \multirow{2}{*}{Momentum \(\alpha\)} & \multicolumn{2}{c}{A$\rightarrow$G} & \multicolumn{2}{c}{G$\rightarrow$A} & \multicolumn{2}{c}{Average} \\
      \cmidrule(r){2-3}\cmidrule(r){4-5}\cmidrule(r){6-7}
      ~ & Rank-1 & mAP & Rank-1 & mAP & Rank-1 & mAP \\
      \midrule
      0.5 & 84.85 & 75.05 & 85.55 & 76.48 & 85.20 & 75.77 \\
      \textbf{0.8 (Ours)} & \textbf{85.16} & \textbf{75.40} & \textbf{85.97} & \textbf{77.02} & \textbf{85.57} & \textbf{76.21} \\
      0.9 & 85.10 & 75.35 & 85.88 & 76.91 & 85.49 & 76.13 \\
      0.99 & 84.65 & 74.80 & 85.32 & 76.18 & 84.99 & 75.49 \\
      \bottomrule
  \end{tabular}
  }
  \label{tab:alpha}
  \vspace{-2mm}
\end{table}
\begin{table}[t]
  \centering
  \caption{Effect of different view imbalance ratios (A:G).}
  \vspace{-2mm}
  \renewcommand{\arraystretch}{1.22}
  \resizebox{0.9\columnwidth}{!}{
  \begin{tabular}{c  c c  c c  c c}
      \toprule
      \multirow{2}{*}{A:G Ratio} & \multicolumn{2}{c}{A$\rightarrow$G} & \multicolumn{2}{c}{G$\rightarrow$A} & \multicolumn{2}{c}{Average} \\
      \cmidrule(r){2-3}\cmidrule(r){4-5}\cmidrule(r){6-7}
      ~ & Rank-1 & mAP & Rank-1 & mAP & Rank-1 & mAP \\
      \midrule
      1:0.25 & 84.68 & 74.82 & 85.20 & 76.30 & 84.94 & 75.56 \\
      1:0.50 & 84.90 & 75.08 & 85.58 & 76.62 & 85.24 & 75.85 \\
      1:0.75 & 85.05 & 75.28 & 85.82 & 76.88 & 85.44 & 76.08 \\
      \textbf{1:1 (Ours)} & \textbf{85.16} & \textbf{75.40} & \textbf{85.97} & \textbf{77.02} & \textbf{85.57} & \textbf{76.21} \\
      \bottomrule
  \end{tabular}
  }
  \label{tab:imbalance}
  \vspace{-2mm}
\end{table}

\begin{table}[t]
  \centering
  \vspace{-2mm}
  \caption{Comparison with different VRD mechanisms.}
  \renewcommand{\arraystretch}{1.22}
  \resizebox{0.9\columnwidth}{!}{
  \begin{tabular}{c  c c c  c c c}
      \toprule
      \multirow{2}{*}{Method} & \multicolumn{3}{c}{A→G} & \multicolumn{3}{c}{G→A} \\
      \cmidrule(r){2-4}  \cmidrule(r){5-7}
       ~ & Rank-1 & mAP & mINP & Rank-1 & mAP & mINP \\
      \midrule
      Down-Conv & 85.16 & 75.40 & 51.35 & 85.97 & 77.02 & 50.42 \\
      Pooling & 85.35 & 75.24 & 51.16 & 85.24 & 77.00 & 50.38 \\
      Projection & 84.88 & 74.79 & 50.27 & 85.55 & 76.92 & 50.08\\
      \bottomrule
  \end{tabular}
  }
  \label{suptab:struct}
\end{table}
\begin{figure}[t]
\centering
\includegraphics[width=1\linewidth]{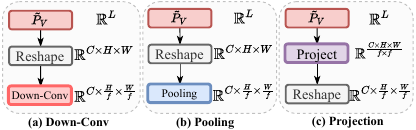}
\caption{Detailed structures of different VRD mechanisms.}
\label{supfig:struct}
\vspace{-2mm}
\end{figure}

\textbf{Effect of Different VRD Mechanisms.}
\newcref{suptab:struct} compares the performance of different VRD mechanisms, with details illustrated in \newcref{supfig:struct}.
The pooling strategy outperforms the projection strategy, likely due to the structural mismatch: UNet processes two-dimensional data, whereas the input viewpoint conditions are one-dimensional.
Pooling at the spatial scale effectively alleviates this issue.
When comparing pooling with convolution, pooling achieves higher Rank-1 under the A→G protocol, but convolution delivers the best overall performance.
This is because pooling excessively compresses viewpoint features, whereas convolution better preserves spatial details.

\begin{figure}[t]
\centering
\includegraphics[width=0.9\linewidth]{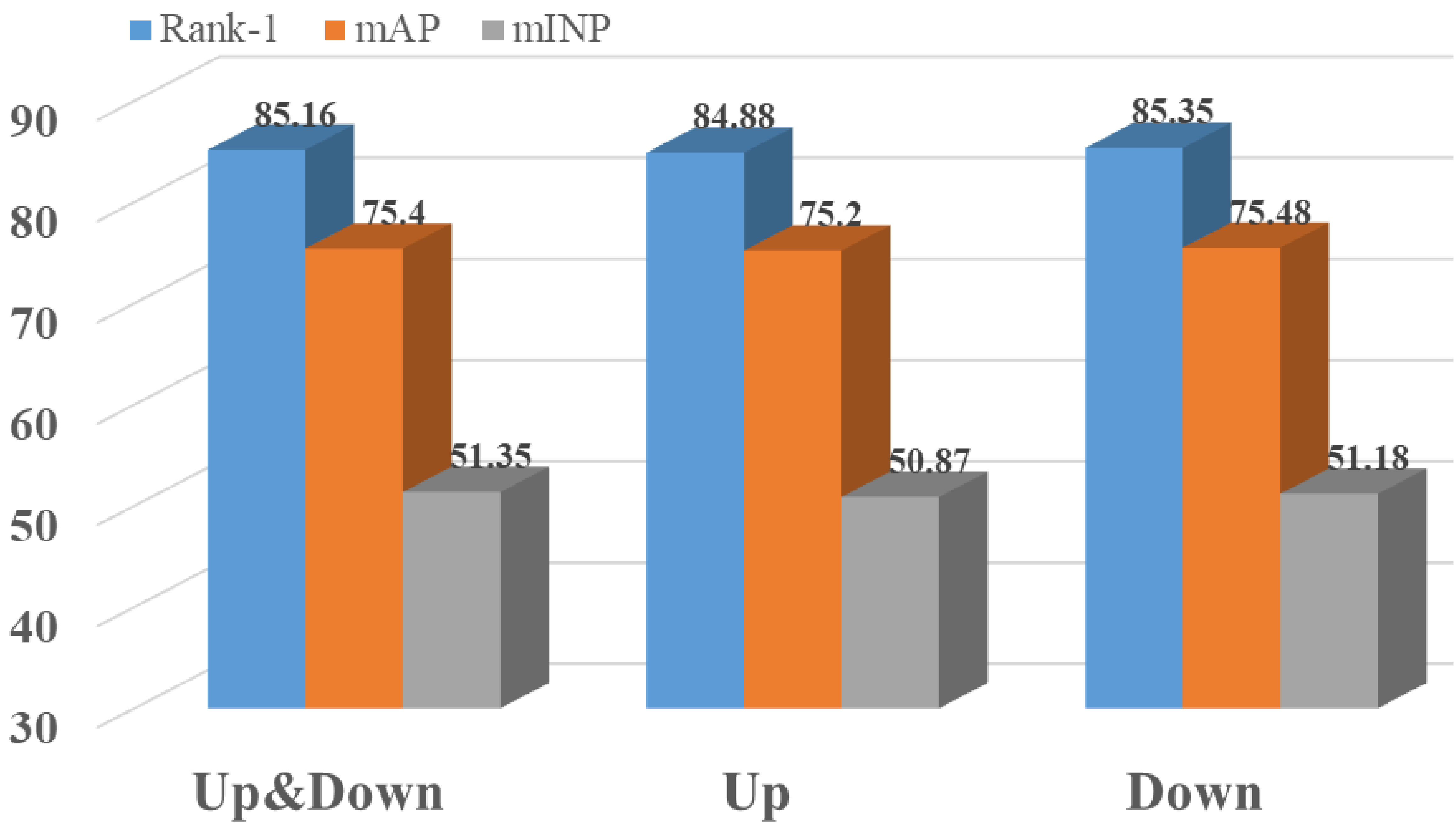}
\caption{
Performance comparison with different insertion positions of VRD under A→G protocol.}
\label{supfig:VRDLoc}
\end{figure}

\textbf{Effect of VRD at Different Positions.}
As described in \newcref{sec:Method}, the VRD module is a plug-and-play component designed to reduce the discrepancy between instance-level and global-level viewpoint conditions.
\newcref{supfig:VRDLoc} reports results when integrating VRD at different positions within the UNet: up-sampling blocks only, down-sampling blocks only, and both.
The performance shows only minor fluctuations across these configurations, with marginal overall differences.
The best results are achieved when VRD is applied in both blocks, which is the default setting in our experiments.
\begin{table}[t]
  \centering
  \caption{Model complexity and inference cost comparison.}
  \vspace{-2mm}
  \renewcommand{\arraystretch}{1.22}
  \resizebox{1\columnwidth}{!}{
  \begin{tabular}{c  c c c c c}
      \toprule
      Method & Params & Train. & GFLOPs & Latency & Thput. \\
      ~ & (M) & (M) & /img & (ms/img) & (img/s) \\
      \midrule
      ViT & 85.75 & 85.747 & 22.682 & 1.487 & 672.43 \\
      Stage I & 85.753 & 85.750 & 22.697 & 1.556 & 642.63 \\
      \textbf{Stage II + SD} & \textbf{952.563} & \textbf{17.610} & \textbf{677.873} & \textbf{26.021} & \textbf{38.43} \\
      \bottomrule
  \end{tabular}
  }
  \label{tab:complexity}
  \vspace{-2mm}
\end{table}
\begin{figure}[t]
\centering
\includegraphics[width=1\linewidth]{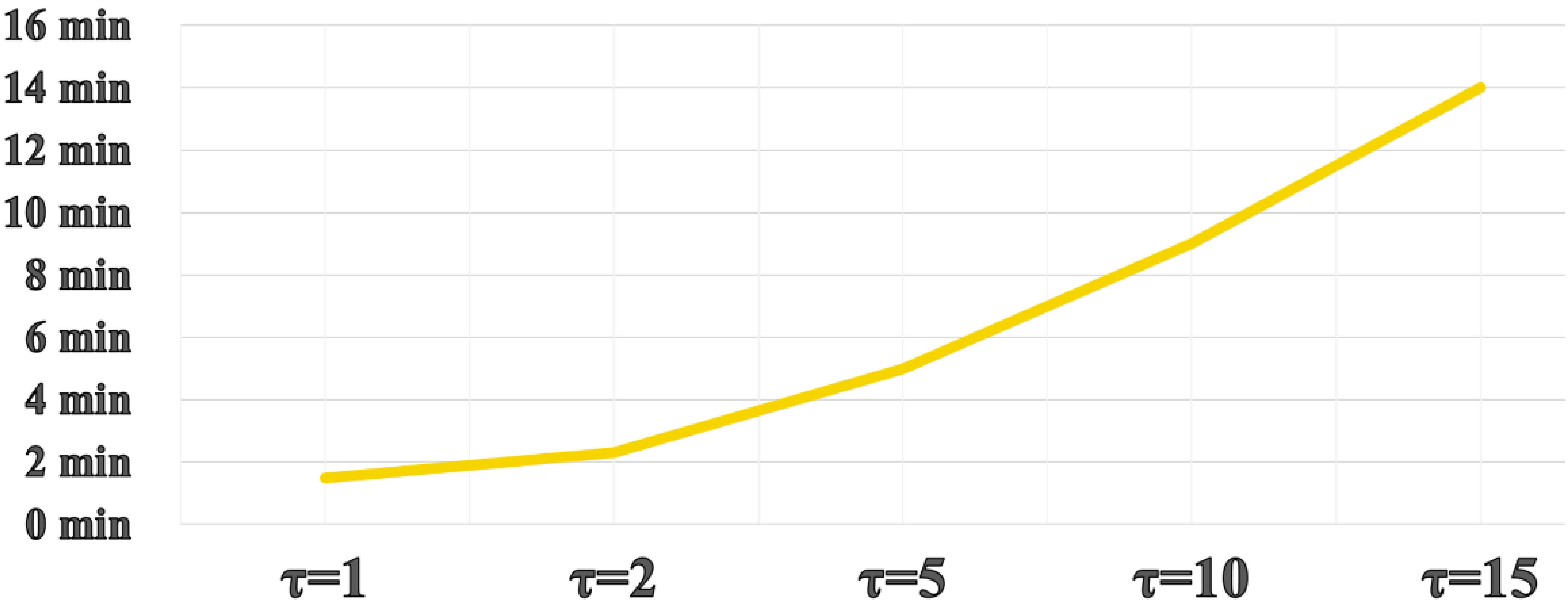}
\caption{Inference time with different timesteps \( \tau \) under the G→A protocol.}
\label{fig:step_time}
\end{figure}
\begin{figure}[t]
\centering
\includegraphics[width=1.0\linewidth]{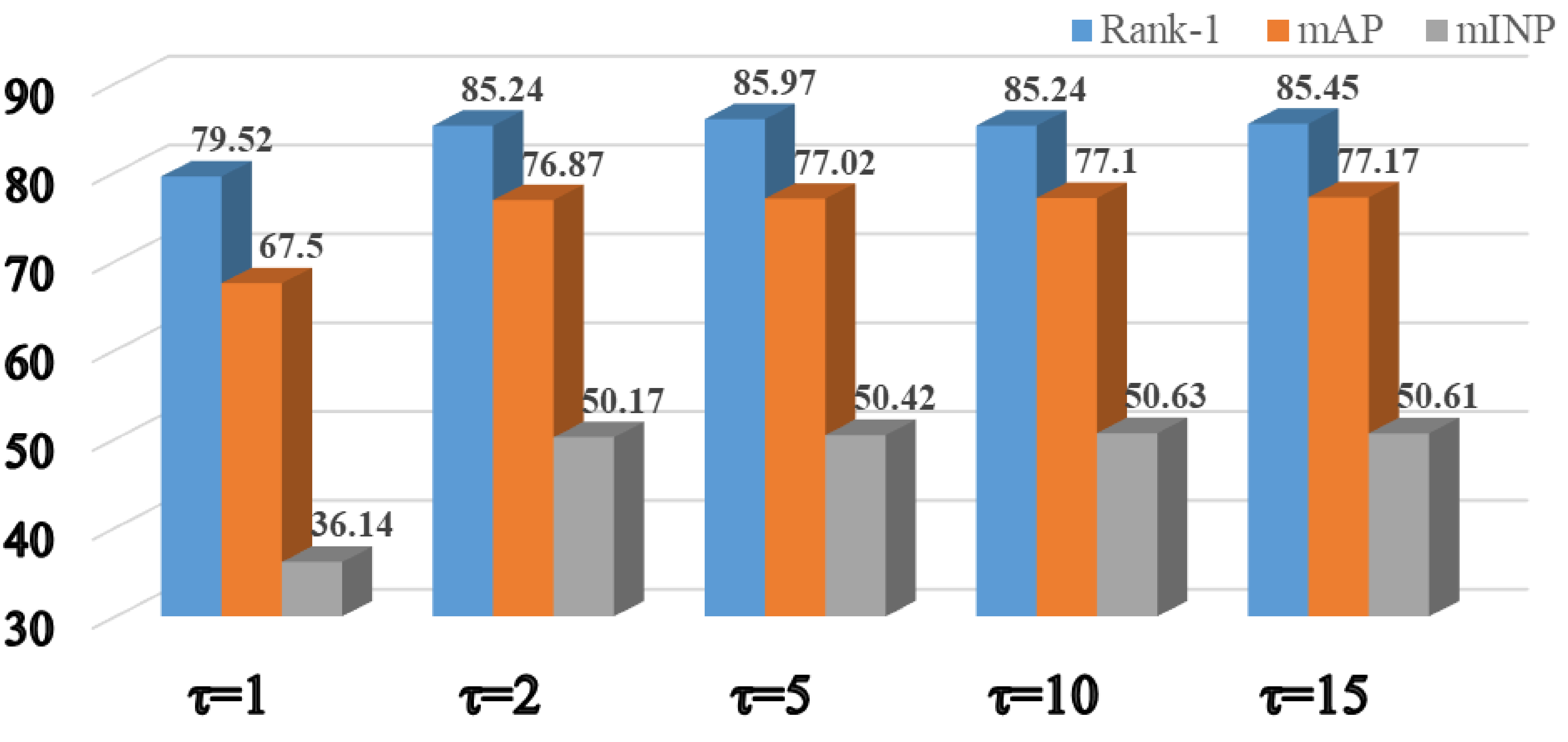}
\caption{Performance with different timesteps \( \tau \) under the G→A protocol.}
\label{fig:inferencestep}
\end{figure}
\begin{figure*}[t]
  \centering
      \includegraphics[width=0.9\linewidth]{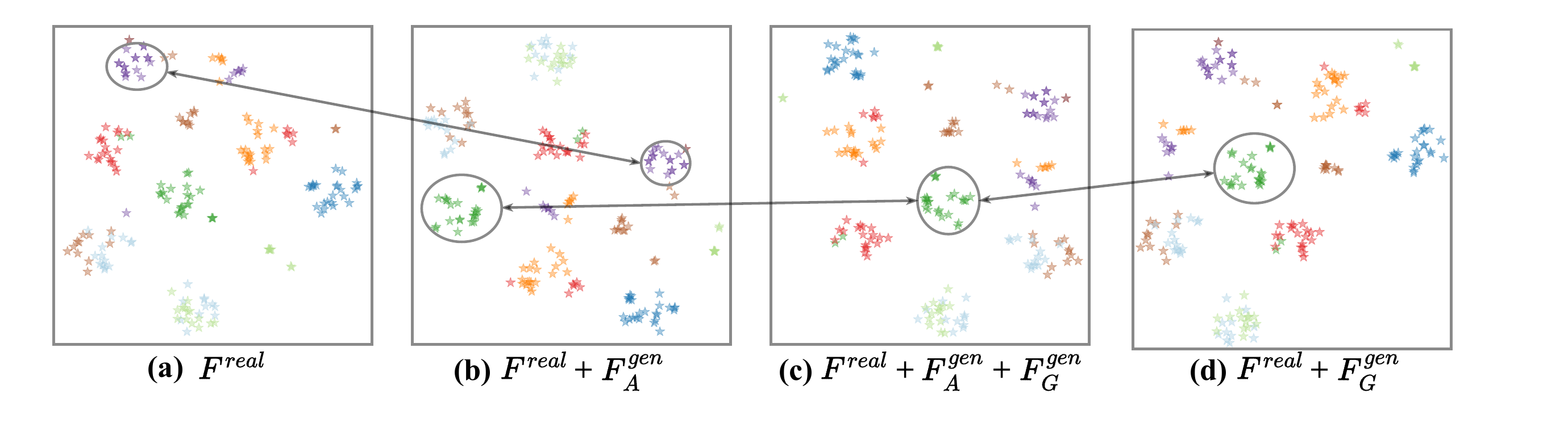}
      \vspace{-2mm}
      \caption{Comparison of feature distributions with t-SNE~\cite{van2008visualizing}. Different colors represent different identities.}
      \label{fig:tsne}
       \vspace{-2mm}
\end{figure*}
\begin{figure}[t]
  \centering
      \includegraphics[width=1\linewidth]{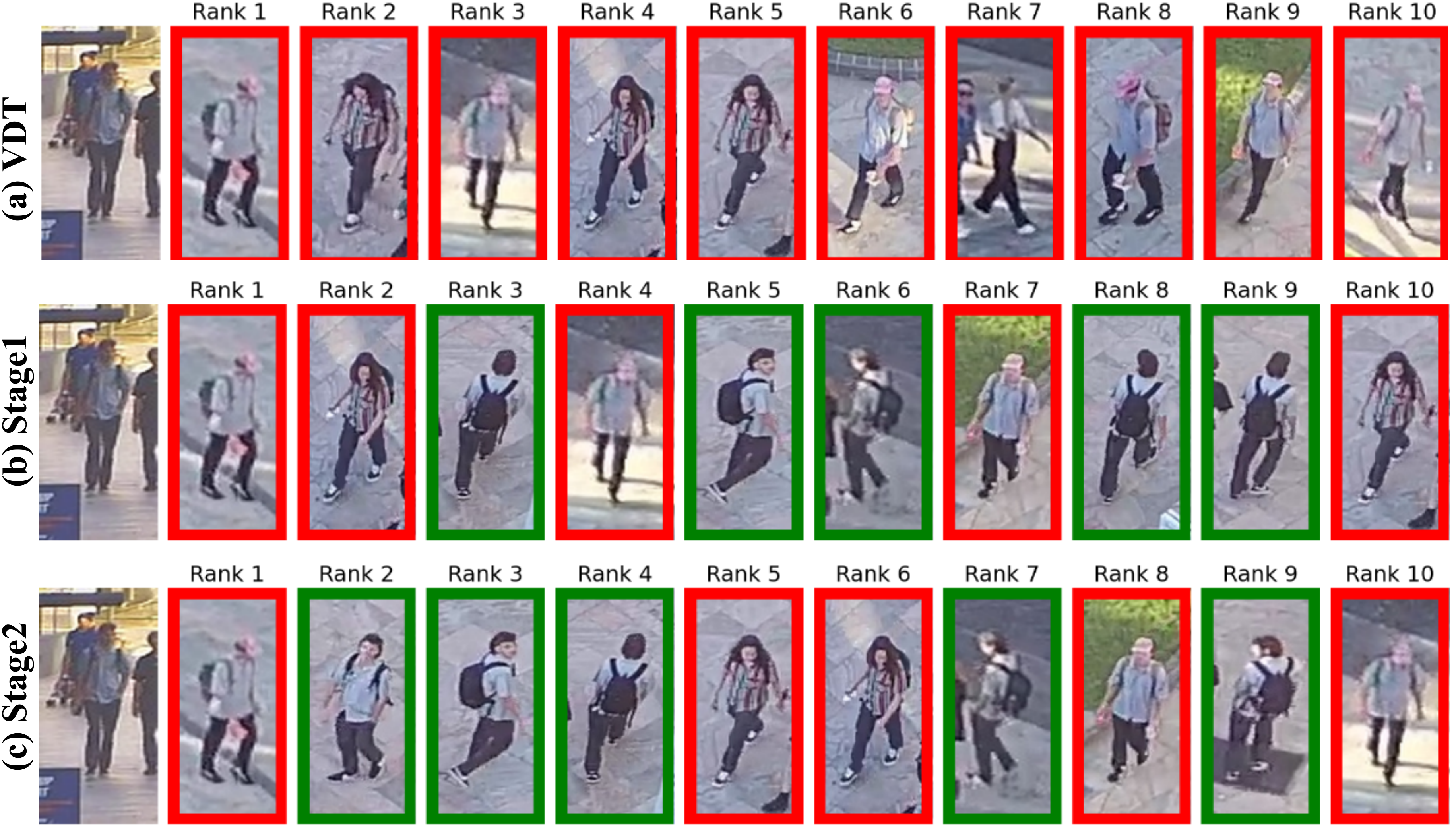}
      \caption{Rank list comparison among VDT, SD-ReID's Stage~I, and SD-ReID's Stage~II on challenging examples. Green boxes indicate correct matches, while red boxes denote incorrect matches.}
      \label{fig:rank}
      \vspace{-3mm}
\end{figure}

\textbf{Effect of Inference Timesteps.}
We analyze the impact of inference timesteps \( \tau \), a key parameter in the denoising process.
Large values incur high computational cost and may even degrade performance, whereas small values result in coarse denoising and low-quality representations.
Since \( \tau \) is the primary source of computational overhead in SD-ReID, we investigate the trade-off between efficiency and accuracy.
As shown in \newcref{fig:step_time} and \newcref{fig:inferencestep}, the inference time grows linearly with \( \tau \), while the retrieval accuracy stabilizes beyond a certain point.
Based on this analysis, we set \( \tau = 5 \), which offers a favorable balance between performance and efficiency.

\textbf{Complexity Analysis.}
\newcref{tab:complexity} reports the model complexity.
The ViT backbone and Stage~I have comparable parameters and latency, confirming that the additional view token introduces negligible overhead.
Incorporating the frozen SD generator in Stage~II raises the total parameter count to 952.56M and the computation to 677.87 GFLOPs/image, yet only 17.61M parameters are trainable.
The resulting latency is 26.02~ms/image, substantially faster than pixel-space diffusion~\cite{rombach2022high} that typically exceeds 1~s/image.
Overall, our proposed SD-ReID remains practical for real-world deployment given the significant performance gains over existing methods.

\subsection{Visualization Analysis}
\label{sec:vis}
\textbf{Feature Distributions.}
\newcref{fig:tsne} visualizes retrieval feature distributions under different settings.
Comparing \newcref{fig:tsne}(a) and (b), generated features \( F^{gen}_A \) produce more compact clusters for distinct identities.
Incorporating features generated under both aerial (\( F^{gen}_A \)) and ground (\( F^{gen}_G \)) views, as shown in \newcref{fig:tsne}(c), further amplifies inter-identity gaps.
Additionally, \newcref{fig:tsne}(c) forms more condensed clusters than \newcref{fig:tsne}(d).
These results demonstrate the effectiveness of leveraging view-specific features to obtain more discriminative features.

\begin{figure}[t]
  \centering
      \includegraphics[width=0.94\linewidth]{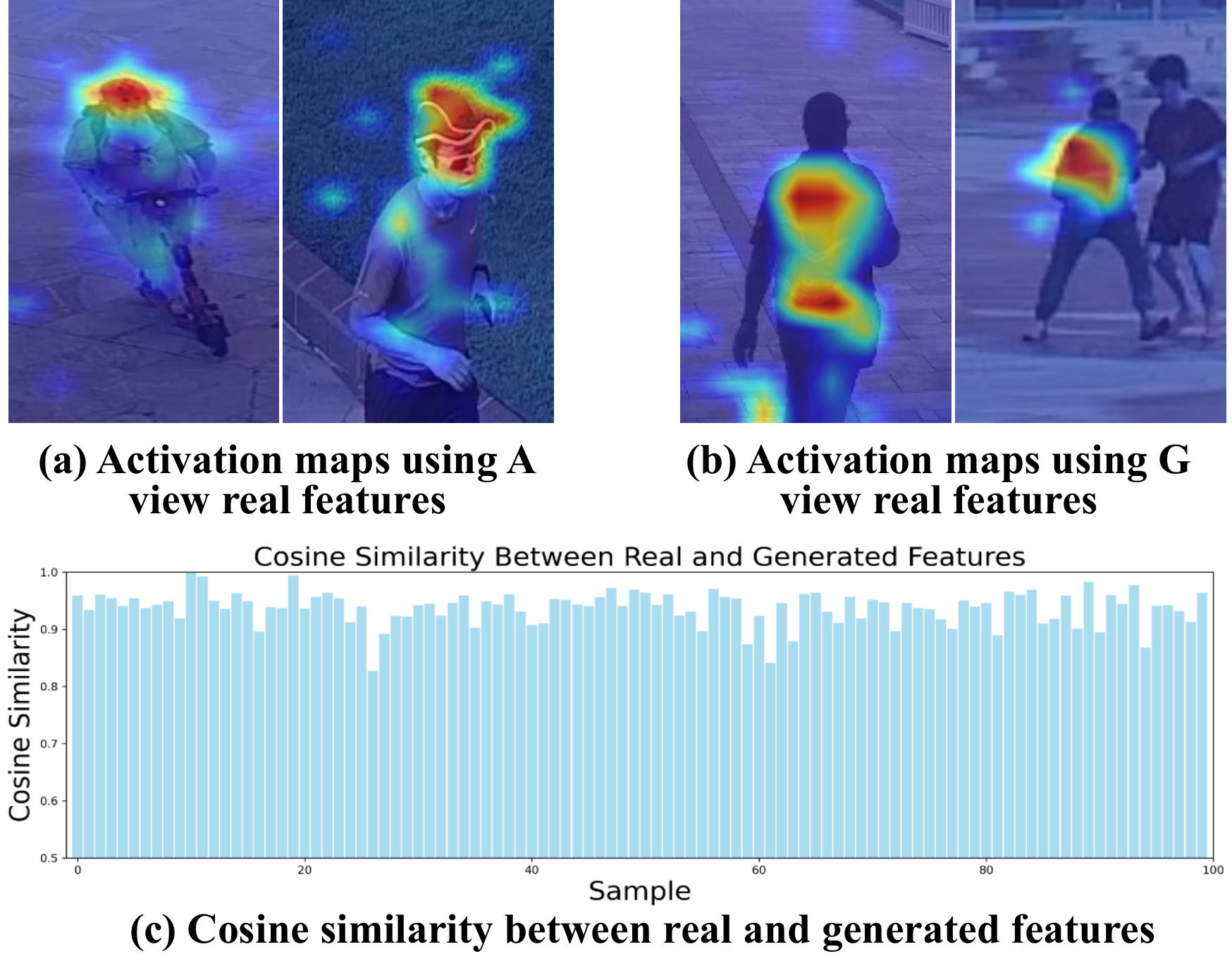}
      \caption{
        Visualization of activation maps and feature similarities. (a) and (b) show activation maps for real aerial and ground images, respectively. (c) shows cosine similarities between real and generated features.}
      \label{fig:viewvis}
      \vspace{-1mm}
\end{figure}
\textbf{Rank List Comparison.}
\newcref{fig:rank} visualizes and compares the rank lists of VDT and the two training stages of SD-ReID on challenging examples from the AG-ReID.v1 dataset.
In \newcref{fig:rank}(a), VDT struggles to correctly identify instances due to the difficulty of directly extracting view-invariant features.
The first training stage, using a simple view-aware discriminative model, substantially improves the matching accuracy.
Integrating the generative model in the second stage further reduces incorrect matches.
These results highlight the effectiveness of SD-ReID in handling challenging retrieval scenarios.

\textbf{Capability of Capturing View-specific Features.}
As shown in \newcref{fig:viewvis}(a) and (b), activation maps reveal view-specific information.
Meanwhile, we compute the cosine similarity between real and generated features in \newcref{fig:viewvis}(c), which demonstrates that the generated features closely resemble the real ones.
These results confirm that generated features effectively capture view-specific information.
\section{Conclusion}
\label{sec:Conclusion}
In this paper, we present SD-ReID, a novel two-stage framework for AG-ReID that leverages generative models to obtain view-specific features, thereby enhancing view-invariant representations.
In the first stage, a view-aware ReID model extracts coarse person representations along with identity and view conditions.
In the second stage, these representations serve as generative targets for a SD model, enabling the generation of view-specific features.
To address the absence of instance-level view conditions during inference, we introduce a memory bank for global-level view conditions and a View-Refined Decoder (VRD) to align generated features with visual features from the ReID backbone, mitigating the distribution gap.
Finally, the refined all-view features are fused with the original ReID features, producing robust representations for retrieval.
Extensive experiments on five AG-ReID benchmarks fully demonstrate the effectiveness of our proposed method.

\bibliographystyle{IEEEtran}
\bibliography{reference}

\end{document}